\documentclass{article}

\usepackage{arxiv}

\usepackage[utf8]{inputenc} 
\usepackage[T1]{fontenc}    
\usepackage{hyperref}       
\usepackage{url}            
\usepackage{booktabs}       
\usepackage{amsfonts}       
\usepackage{nicefrac}       
\usepackage{microtype}      
\usepackage{lipsum}		
\usepackage{graphicx}
\usepackage{natbib}
\usepackage{doi}
\usepackage{amsmath}

\usepackage{tabularx}

\title{vi-mistral-x}

\author{ \href{https://orcid.org/0000-0000-0000-0000}{\includegraphics[scale=0.06]{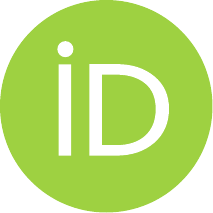}\hspace{1mm}James Vo}\thanks{http://agilesoda.ai} \\
	AI Algorithm Research Team\\
	AgileSoDA Inc.\\
	Seoul, South Korea \\
	\texttt{anhdungitvn@agilesoda.ai} \\
}

\renewcommand{\headeright}{Technical Report}
\renewcommand{\undertitle}{Building a Vietnamese Language Model with Advanced Continual Pre-training}

\hypersetup{
pdftitle={vi-mistral-x},
pdfsubject={q-bio.NC, q-bio.QM},
pdfauthor={David S.~Hippocampus},
pdfkeywords={Vietnamese, LLM, CPT},
}

\begin{document}
\maketitle

\begin{abstract}
	The advancement of Large Language Models (LLMs) has significantly transformed the field of natural language processing, although the focus on English-centric models has created a noticeable research gap for specific languages, including Vietnamese. To address this issue, this paper presents vi-mistral-x, an innovative Large Language Model designed expressly for the Vietnamese language. It utilizes a unique method of continual pre-training, based on the Mistral architecture, which incorporates grouped-query attention and sliding window attention techniques. This model, vi-Mistral-X, marks a significant step forward in improving the understanding and generation of the Vietnamese language. It introduces an additional phase of continual pre-training, specifically adapted for Vietnamese, enhancing the model's capability in understanding complex language nuances and generating accurate, context-aware Vietnamese text. Through comprehensive testing on various benchmarks, vi-mistral-x has shown to outperform existing Vietnamese LLMs in several key areas, including text classification, question answering, and text generation. Particularly, in the Vietnamese Multitask Language Understanding (VMLU) benchmark, vi-mistral-x sets a new standard, outperforming other available models significantly.  This paper highlights the critical role of continual pre-training in advancing language-specific LLMs and opens new avenues for the development of multilingual models. We aim for vi-mistral-x to not just be an important asset for processing the Vietnamese language but also to encourage more advancements in creating large language models for languages that are less represented.
\end{abstract}


\section{Introduction}
The field of natural language processing (NLP) has witnessed a paradigm shift with the advent of Large Language Models (LLMs), which have shown tremendous potential in understanding and generating human language. LLMs like ChatGPT and GPT-4 have paved the way for innovations that edge closer to achieving Artificial General Intelligence (AGI). However, the progress in this domain has predominantly centered around English, leading to a substantial disparity in the development and performance of LLMs for other languages. This disparity not only limits the global applicability of such models but also underscores a crucial gap in the research and development of language models that cater to the diverse linguistic landscape of our world.

In particular, the Vietnamese language, with its unique syntactic and semantic complexities, has not been adequately represented in the current wave of LLM advancements. This oversight hinders the ability of Vietnamese NLP applications to achieve the same level of sophistication and effectiveness as their English counterparts, thereby creating a significant bottleneck in the development of Vietnamese language technology.

To bridge this gap, our paper introduces vi-mistral-x, an LLM specifically designed to address the challenges associated with processing and generating the Vietnamese language. Building on the foundation laid by the innovative Mistral architecture, vi-mistral-x incorporates advanced techniques such as grouped-query attention (GQA) and sliding window attention (SWA) \cite{jiang2023mistral}. These features are part of a unique approach to continual pre-training that is tailored to the Vietnamese language, enabling the model to capture its linguistic nuances more accurately.

The development of vi-mistral-x is inspired by recent efforts in the field to extend the capabilities of existing models to additional languages. This includes the adaptation of LLaMA for the Chinese language \cite{cui2024efficient} and the Korean language \cite{l._junbum_2023}. By employing a similar methodology of extending vocabulary and incorporating language-specific pre-training and fine-tuning phases, we aim to achieve a leap in the quality of text understanding and generation in Vietnamese. This approach is underscored by the success of the Mistral 7B model, which has demonstrated the effectiveness of GQA and SWA in improving performance and efficiency across various NLP tasks.

Our work on vi-mistral-x represents a critical step toward closing the research gap for the Vietnamese language within the NLP community. By detailing our approach and sharing our findings, we hope to not only enhance the capabilities of language models for Vietnamese but also to encourage further research and development efforts focused on other underrepresented languages. This endeavor aligns with our broader goal of promoting inclusivity and diversity in the advancement of natural language processing technologies, ensuring that the benefits of these innovations are accessible to a wider audience around the globe.

\section{Proposed Method}
\label{sec:headings}

This section details the methodology employed in developing vi-Mistral-X, focusing on the adaptation of the Mistral architecture for the Vietnamese language. The process encompasses five main stages: corpus preparation, tokenizer training, model initialization, model training, and model alignment.

\subsection{Effective Corpus Preparation}
The stage involves refining a Vietnamese text corpus extracted from CulturaX, a comprehensive multilingual dataset designed to support the development of Large Language Models (LLMs) across 167 languages, including Vietnamese~\cite{nguyen2023culturax}. The primary goal was to reduce the original corpus to a more manageable size while enhancing its quality, which is critical for the effective training of language models. We employed a multi-step preprocessing pipeline with the following components:

\paragraph{Random Selection}
As an initial step, we used a random selection technique to significantly reduce the corpus's size. This method allowed us to manage computational resources better by focusing on a smaller, yet representative, subset of the original dataset.

\paragraph{N-gram-based Filtering for Deduplication}
We applied an n-gram-based filtering method to ensure the dataset's uniqueness. This technique analyzes the frequency of contiguous sequences of n-gram in the text to identify and remove duplicate or nearly identical content. Such deduplication is crucial to reduce the risk of overfitting the model on repetitive data.

\paragraph{BERT-based Binary Classifier for Toxicity Filtering}
To further enhance the corpus quality, we used a high-precision BERT-based binary classifier to filter out toxic content. Deploying this classifier helped exclude data that could propagate undesirable biases or harmful expressions within the trained model.

\paragraph{Perplexity-based Filtering}
The final preprocessing step was perplexity-based filtering. Perplexity, a measure of a probability model's predictive accuracy, was used to assess and filter documents based on their coherence and quality. This criterion is vital for developing language models, as it ensures that only high-quality and coherent documents contribute to the training process.

This comprehensive preprocessing pipeline was designed to enhance the quality of the Vietnamese corpus from CulturaX. Table 1 presents a detailed comparison between the original CulturaX corpus and the refined corpus used for training the vi-mistral-x model. Although a formal evaluation employing quantitative measures to ascertain the processed data's specific impact on model training compared to the original data has not yet been conducted, there is reason to believe that selecting and refining data to improve the consistency and quality of each data sample can lead to enhanced computational efficiency. Specifically, reducing the size of the data by removing noisy and non-uniform samples can decrease computing costs due to lower resource requirements and may also improve the training quality of the model by concentrating on high-quality data, thereby optimizing the learning process.

\begin{table}[ht]
\centering
\begin{tabular}{lrr}
\hline
                       & CulturaX/vi & Selected corpus \\ \hline
No. of documents       & 54,988,654  & 7,331,840       \\
Size in GB (parquet)   & 150.91      & 20.656          \\
No. of tokens          & NA          & 8,323,137,536   \\ \hline
\end{tabular}
\caption{Detailed comparison of the original CulturaX/vi and the refined corpus for vi-mistral-x model training}
\label{your-table-label}
\end{table}

\subsection{Effective Tokenizer Training}

The second phase in the adaptation of the pretrained Mistral model for Vietnamese language processing involves the development of a tokenizer capable of efficiently handling Vietnamese text. Initially, we utilized Google SentencePiece\footnote{https://github.com/google/sentencepiece} to train a new SentencePiece model (SPM). Subsequently, we performed rule-based token filtering on the trained SPM, with a focus on Vietnamese character recognition. The enhanced SPM was then integrated with the original Mistral's SPM model. This hybrid tokenizer maintains the ability to process English and other languages previously supported by Mistral-7B, while also effectively managing Vietnamese text. This capability is pivotal for facilitating bilingual or multilingual continual training in the future.

\paragraph{SPM Model Training}
The new SPM model was developed by employing Google SentencePiece to train a new model on our refined corpus, which was obtained in the initial stages. The corpus was significantly reduced to a manageable size (20GB) without necessitating extra sampling, limiting maximum sentence length, or filtering character coverage. The vocabulary size was determined by balancing the trade-off between input complexity and model complexity. A larger vocabulary tends to decrease the number of tokens passed to the model, thereby reducing input complexity, but it increases model complexity due to the expansion of the embedding and language model head dimensions. As illustrated in Figure 1, a vocabulary size of 8,096 is optimal for our dataset and the Mistral model, based on our observations.

\paragraph{SPM Model Refining}
This phase involved the removal of abnormal characters from the trained SPM model to achieve a high-quality and coherent tokenizer. The refinement rules were established based on manual definitions and prioritized tokens with the highest frequency.

\paragraph{Model Combination}
The refined SPM was integrated with the original Mistral's SPM model to create the final tokenizer. This integration process involved several rounds of tokenizer training and analysis, ensuring the new tokenizer model includes a comprehensive and relevant vocabulary for our project.

This meticulous approach to tokenizer training and refinement underscores the importance of adapting language processing tools to efficiently manage specific linguistic characteristics, thereby enhancing bilingual or multilingual training capabilities.

\begin{figure}[h] 
    \centering
    \includegraphics[width=0.8\textwidth, height=0.3\textheight, keepaspectratio]{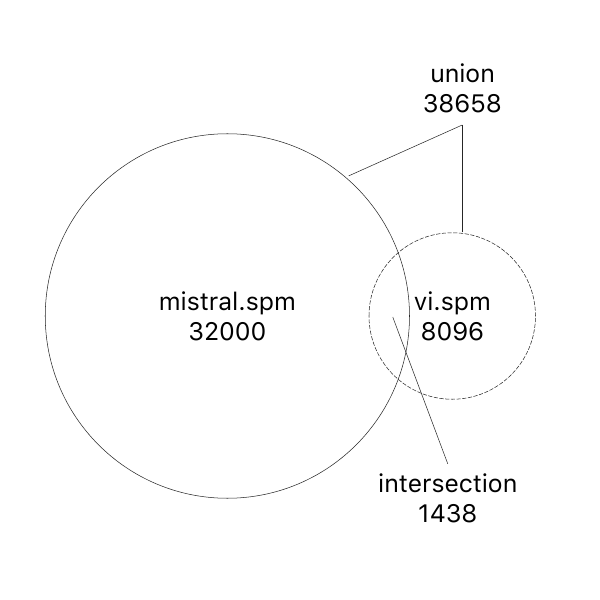}
    \caption{Visualization of the Mistral SPM model and customized Vietnamese SPM}
    \label{fig:yourlabel}
\end{figure}

\begin{table}[ht]
\centering
\begin{tabular}{ccc}
\hline
\textbf{Vocab Size} & \textbf{Relative Input Complexity} & \textbf{Relative Model Embedding Complexity} \\ \hline
1000 & 0.841395049 & 1.01640625 \\
2000 & 0.630757254 & 1.04403125 \\
3000 & 0.584167181 & 1.073125 \\
4000 & 0.557951001 & 1.1025 \\
5000 & 0.539983642 & 1.13171875 \\
6000 & 0.526697283 & 1.16078125 \\
7000 & 0.516199365 & 1.18978125 \\
8000 & 0.50747301 & 1.21909375 \\
9000 & 0.500297369 & 1.2479375 \\
10000 & 0.494061494 & 1.27684375 \\
11000 & 0.488440053 & 1.30575 \\
12000 & 0.483667799 & 1.33428125 \\
13000 & 0.479315439 & 1.36334375 \\
14000 & 0.475368937 & 1.39240625 \\
15000 & 0.471816595 & 1.42109375 \\
16000 & 0.468521178 & 1.4500625 \\
17000 & 0.465535533 & 1.4789375 \\
18000 & 0.462752376 & 1.50753125 \\
19000 & 0.460176296 & 1.5363125 \\
20000 & 0.45770208 & 1.5655625 \\
30000 & 0.439826305 & 1.85509375 \\
40000 & 0.422321581 & 2.14471875 \\
80000 & 0.403156539 & 3.31371875 \\
120000 & 0.395356195 & 4.49840625 \\ \hline
\end{tabular}
\caption{Input Complexity and Model Embedding Complexity by Vocab Size}
\label{table:vocab_complexity}
\end{table}

\begin{figure}[htbp]
\centering
\includegraphics[width=0.8\textwidth]{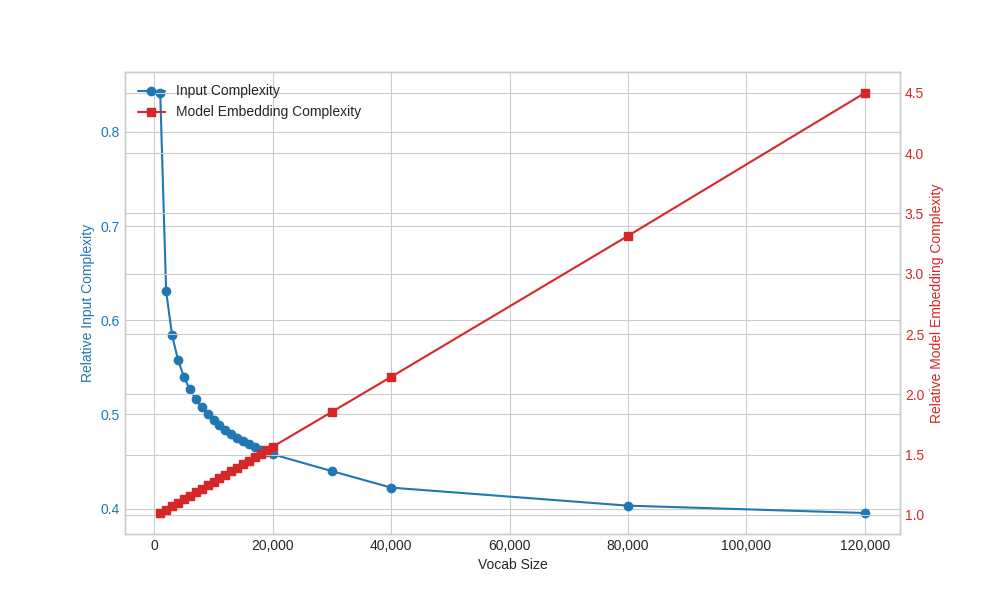}
\caption{Relative Input Complexity and Relative Model Embedding Complexity by Vocab Size.}
\label{fig:vocab_complexity}
\end{figure}

\subsection{Effective Model Initialization}
For model initialization, we adapted the Mistral architecture to accommodate the newly-generated Vietnamese token embeddings produced by the novel tokenizer. This adaptation necessitated the expansion of both the model’s embedding layer and language model head to include the Vietnamese-specific tokens, whilst preserving the integrity of the original model's architecture. Figure 2 illustrates the architectural comparison between the original Mistral framework and our modified version tailored to accommodate Vietnamese-specific tokens.

\begin{figure}[h] 
    \centering
    \includegraphics[width=0.8\textwidth, height=0.4\textheight, keepaspectratio]{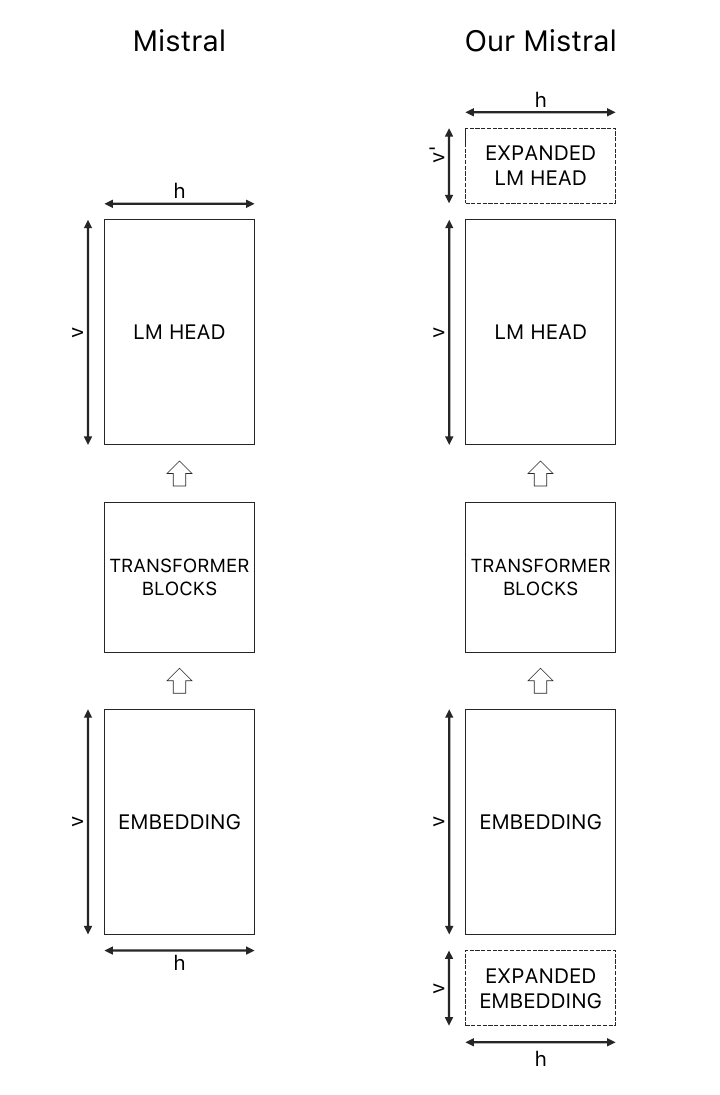}
    \caption{Model Architecture of the Mistral Model and Our Expanded Model}
    \label{fig:yourlabel}
\end{figure}

\paragraph{Initilization}

Let \(V = \{1,\ldots, n\}\) be the model's vocabulary, where \(n=32000\).
Let \(w_{1:T}\) be a sequence of words.
Let \(p_{\theta}(w_{i}|w_{1:i-1})\) be the LM parameterized by \(\theta\), defined by:

\[
P(\boldsymbol{w}_i | \boldsymbol{w}_{i-n+1:i-1}) = \frac{\exp(\boldsymbol{h}_{i-1}^T \boldsymbol{e}_{w_i})}{\sum_{j=1}^{m} \exp(\boldsymbol{h}_{i-1}^T \boldsymbol{e}_j)}
\]

where \(\boldsymbol{h}_{i-1}=\phi_{\theta}(w_{1:i-1}) \in \mathbb{R}^{d}\) is the representation of the prefix, and \(\boldsymbol{e}_{i} \in \mathbb{R}^{d}\) is the embedding for word \(i \in V\). The \(\boldsymbol{e}_{i}\) are contained in \(\theta\).

When a new word, \(\text{input\_ids} \in [32000, 38658]\), \(n+1 \notin V\) is added to the vocab of the pretrained LM, the new word \(n+1 \notin V\)'s embedding needs to be initialized as \(\boldsymbol{e}_{n+1}\).
Let \(p_{\theta'}(w_{i}|w_{1:i-1})\) be the new LM, which has parameters \(\theta'=\theta \cup \{\boldsymbol{e}_{n+1}\}\), defined by:

\[
p_{\theta'}(w_i | w_{1:i-1}) = \frac{\exp(\boldsymbol{h}_{i-1}^T \boldsymbol{e}_{w_i})}{\sum_{j=1}^{m} \exp(\boldsymbol{h}_{i-1}^T \boldsymbol{e}_j) + \exp(\boldsymbol{h}_{i-1}^T \boldsymbol{e}_{n+1})}
\]

\[
p_{\theta}(w_i | w_{1:i-1}) = p_{\theta}(w_i | w_{1:i-1}) \times \frac{1}{1 + \exp(\boldsymbol{h}_{i-1}^T \boldsymbol{e}_{n+1})/\sum_{j=1}^{m} \exp(\boldsymbol{h}_{i-1}^T \boldsymbol{e}_j)}
\]

The updated probability of a particular word is the original probability of that word scaled by a multiplicative element that is less than one, denoted as \(\frac{1}{1 + \exp(\boldsymbol{h}_{i-1}^T \boldsymbol{e}_{n+1})/\sum_{j=1}^{m} \exp(\boldsymbol{h}_{i-1}^T \boldsymbol{e}_j)}\).
This leads the model to incorrectly assign a probability of 1 to newly added words and 0 to the original words.

Therefore, the new embedding is adjusted as follows:

\[
\exp(\boldsymbol{h}_{i}^T \boldsymbol{e}_{n+1}) = \exp\left(\boldsymbol{h}_{i}^T \frac{1}{n} \sum_{j=1}^{n} \boldsymbol{e}_{j}\right)
\]

Finally, the model's embedding and language model (LM) head are resized to \(V' \times H\) and \(H \times V'\), respectively, where \(V' = 38659\).

\subsection{Effective Model training}

\paragraph{Memory and Computational Efficiency in Training}
The core of vi-mistral-x’s development involved continual pre-training on a Vietnamese corpus. In this stage, our research is driven by the need to leverage the most effective resources in computational linguistics and machine learning. We focus on addressing two main challenges in Large Language Models (LLMs): memory capacity limitations, which lead to Out Of Memory (OOM) errors, and the requirement for significant computational power, causing long training times. Our work seeks to overcome these issues by optimizing the model architecture and training processes.

In our pursuit, we have concentrated on a curated selection of model architectures, including \href{https://llama.meta.com/}{Llama2}, \href{https://mistral.ai/}{Mistral}, and \href{https://ai.google.dev/gemma}{Gemma}. These were chosen based on their potential for high efficiency and compatibility with our objectives. Additionally, our strategy encompasses the integration of advanced parallelism techniques, such as Fully Sharded Data Parallelism (FSDP), DeepSpeed ZeRO-3 (DSZERO3), Pipeline Parallelism (PP), and Tensor Parallelism (TP). These methods are instrumental in distributing the computational load and memory usage across multiple devices, thereby alleviating the aforementioned constraints.

Our optimizations have significantly increased training speed, making our library about twice as fast as similar open-source options. Specifically, it's 1.6 times faster than both \cite{dao2023flashattention2} and the PyTorch version of \href{https://pytorch.org/tutorials/intermediate/scaled_dot_product_attention_tutorial.html}{Scaled Dot Product Attention (SDPA) (2024)}.

Our findings highlight the possibility of greatly improving the efficiency of training transformer-based models, advancing artificial intelligence research.

\paragraph{Optimization}

Let \(W \notin \mathbb{R}^{m \times n}\) be a weight matrix. Let
\[G_{t} = -\nabla_{W} \phi_t(W_t) \in \mathbb{R}^{m \times n}\]
be the gradient matrix at step \(t\). The updated weight matrix is computed by:
\[W_T = W_0 + \eta \sum_{t=0}^{T-1} G_t\]
where \(\eta\) is the learning rate and \(\phi_t\) is a stateful gradient regularizer, which is memory-intensive. For instance, [AdamW](https://pytorch.org/docs/stable/generated/torch.optim.AdamW.html) takes \(4 \times m \times n\) memory for gradient, variance, momentum, and parameters.

To enhance memory and computational efficiency during training, LoRA and its derivatives, which perform a low-rank projection in the weight space as \(W_T = W_0 + B_T A_T\), were not selected due to their inherent low-rank limitations. Our goal is to achieve a full-rank update that is both memory- and computation-efficient. Therefore, `Tokenizer`, `Model`, `Trainer`, and `Optimizer` are imported from our XLLM library \cite{jamesvo2023xllm}, which has the same interface as those in the [Transformers](https://github.com/huggingface/transformers.git) library. Techniques such as learning rate warm-up and learning rate scheduling, which adjust the learning rate appropriately across layers, are also applied to optimize the training process.

\paragraph{Training}
The model underwent training on a computational framework consisting of eight Nvidia H100 80GB SXM5 GPUs. Due to the intentional interruption of the training process for purposes of advanced evaluation and optimization, an exact duration of training was not documented. However, a rough estimate suggests that, under conditions of uninterrupted training, the process would span approximately 104 hours. Financially, this duration translates to an approximate expenditure of $3902.08, given the operational cost of $37.52 per hour per node.

\subsection{Model alignment}

Following the pre-training phase, vi-mistral-x underwent a series of fine-tuning processes aimed at aligning the model with specific NLP tasks. This alignment involved training the model on task-specific Vietnamese datasets, such as text classification, question answering, and text generation. Each task-focused fine-tuning phase allowed vi-mistral-x to adjust its parameters to optimize performance on that task, thereby ensuring its applicability and effectiveness across a wide range of NLP applications. This step was crucial for benchmarking vi-mistral-x against existing Vietnamese LLMs and demonstrating its superior performance across several key areas.

Through these methodological steps, vi-mistral-x represents a significant advancement in the development of LLMs for the Vietnamese language, offering enhanced understanding and generation capabilities that set a new benchmark for performance in Vietnamese NLP tasks.

\section{Experimental Results}
\label{sec:headings}

\subsection{Pretrained Model}
\subsubsection{Loss and Accuracy}
\paragraph{References}

\begin{itemize}
  \item \href{https://huggingface.co/anhdungitvn/vi-mistral-x}{anhdungitvn/vi-mistral-x}
  \item \href{https://huggingface.co/mistralai/Mistral-7B-v0.1}{mistralai/Mistral-7B-v0.1} \cite{jiang2023mistral}
  \item \href{https://huggingface.co/Viet-Mistral/Vistral-7B-Chat}{Viet-Mistral/Vistral-7B-Chat} \cite{chien2023vistral}
  \item \href{https://huggingface.co/vinai/PhoGPT-7B5}{vinai/PhoGPT-7B5} \cite{nguyen2024phogpt}
  \item \href{https://huggingface.co/bkai-foundation-models/vietnamese-llama2-7b-120GB}{bkai-foundation-models/vietnamese-llama2-7b-120GB}
  \item \href{https://huggingface.co/meta-llama/Llama-2-7b-hf}{meta-llama/Llama-2-7b-hf} \cite{touvron2023llama}
  \item \href{https://huggingface.co/meta-llama/Llama-2-13b-hf}{meta-llama/Llama-2-13b-hf} \cite{touvron2023llama}
  \item \href{https://huggingface.co/google/gemma-7b}{google/gemma-7b} \cite{gemmateam2024gemma}
\end{itemize}

\paragraph{Setting}
\begin{itemize}
  \item Task: CLM (next token prediction)
  \item Test data: \href{https://huggingface.co/datasets/anhdungitvn/wiki_vi_splitted/tree/main}{anhdungitvn/wiki\_vi\_splitted}
  \item Test data selection: random train\_test\_split
  \item Test data size: 10000 documents
  \item Metrics:
  \begin{itemize}
    \item Tokens: smaller is better
    \item Loss: smaller is better
    \item Accuracy: larger is better
  \end{itemize}
\end{itemize}

\paragraph{Experimental Results}

\begin{table}[ht]
\centering
\begin{tabular}{llllll}
\hline
\textbf{Model} & \textbf{Type} & \textbf{Length} & \textbf{Tokens} & \textbf{Loss} & \textbf{Accuracy}***** \\ \hline
\textbf{anhdungitvn/vi-mistral-x*} & \textbf{Pretrained} & \textbf{4096} & \textbf{2068480} & \textbf{2.1566} & \textbf{0.5622} \\
mistralai/Mistral-7B-v0.1 & Pretrained & 4096 & 4517888 & 1.3687 & 0.6813 \\
Viet-Mistral/Vistral-7B-Chat & \textit{Finetuned}** & 4096 & 2224128 & 1.7354 & 0.6223 \\
vinai/PhoGPT-7B5 & Pretrained & \textit{2048***} & 1982464 & 16.5563**** & 0.0029 \\
bkai.../vietnamese-llama2-7b-120GB & Pretrained & 4096 & 2191360 & 2.4808 & 0.5207 \\
meta-llama/Llama-2-7b & Pretrained & 4096 & 4632576 & 1.1287 & 0.7295 \\
meta-llama/Llama-2-13b & Pretrained & 4096 & 4632576 & 0.9543 & 0.7700 \\
google/gemma-7b & Pretrained & 4096 & 2232320 & \textit{...} & \textit{...} \\ \hline
\end{tabular}
\caption{Comparison of Pretrained Models}
\label{tab:language_models}
\end{table}

\begin{itemize}
    \item [*] The model vi-mistral-x* is currently under development. The shown results were obtained by evaluating a checkpoint at epoch 0.08.
    \item[**] The Viet-Mistral/Vistral-7B pretrained model is unpublished, so we evaluated the Viet-Mistral/Vistral-7B finetuned model.
    \item[***] The model \texttt{vinai/PhoGPT-7B5} doesn't support an input sequence length of 4096. A RuntimeError occurs in \href{https://huggingface.co/vinai/PhoGPT-7B5/blob/main/modeling_mpt.py}{modeling\_mpt.py} on line 138: "The size of tensor a (4096) must match the size of tensor b (2048) at non-singleton dimension 3."
     \item[****] The same evaluation method was applied to all models. The results indicate that the loss for this particular model is unusually high, suggesting that the evaluation method employed may not be appropriate for this model. Further investigation is required.
    \item[*****] Improved accuracy in a Causal Language Model (CLM) for next-token prediction does not guarantee enhanced performance in other tasks or on different datasets. Loss and accuracy metrics merely indicate the model's current training state and can differ substantially among various models. Therefore, they cannot be directly compared based solely on loss and accuracy.
\end{itemize}

\subsubsection{Vietnamese Multitask Language Understanding (VMLU)}
\paragraph{References} \mbox{}\\
\begin{itemize}
  \item \href{https://vmlu.ai/}{VMLU}
  \item \href{https://huggingface.co/datasets/anhdungitvn/vmlu_v1.5}{anhdungitvn/vmlu\_v1.5}
\end{itemize}

\paragraph{VMLU}
VMLU is a benchmark suite aimed at evaluating foundation models' capabilities, focusing on the Vietnamese language. It includes 10,880 multiple-choice questions across 58 subjects within STEM, Humanities, Social Sciences, and more, covering difficulty levels from basic to advanced.

\paragraph{Dataset: anhdungitvn/vmlu\_v1.5} \mbox{}\\
The dataset \texttt{anhdungitvn/vmlu\_v1.5} was originally created from \texttt{vmlu\_v1.5} by formatting it into the Hugging Face datasets format for easier use.

\paragraph{Example}

\begin{figure}[h] 
    \centering
    \includegraphics[width=0.8\textwidth, height=0.35\textheight, keepaspectratio]{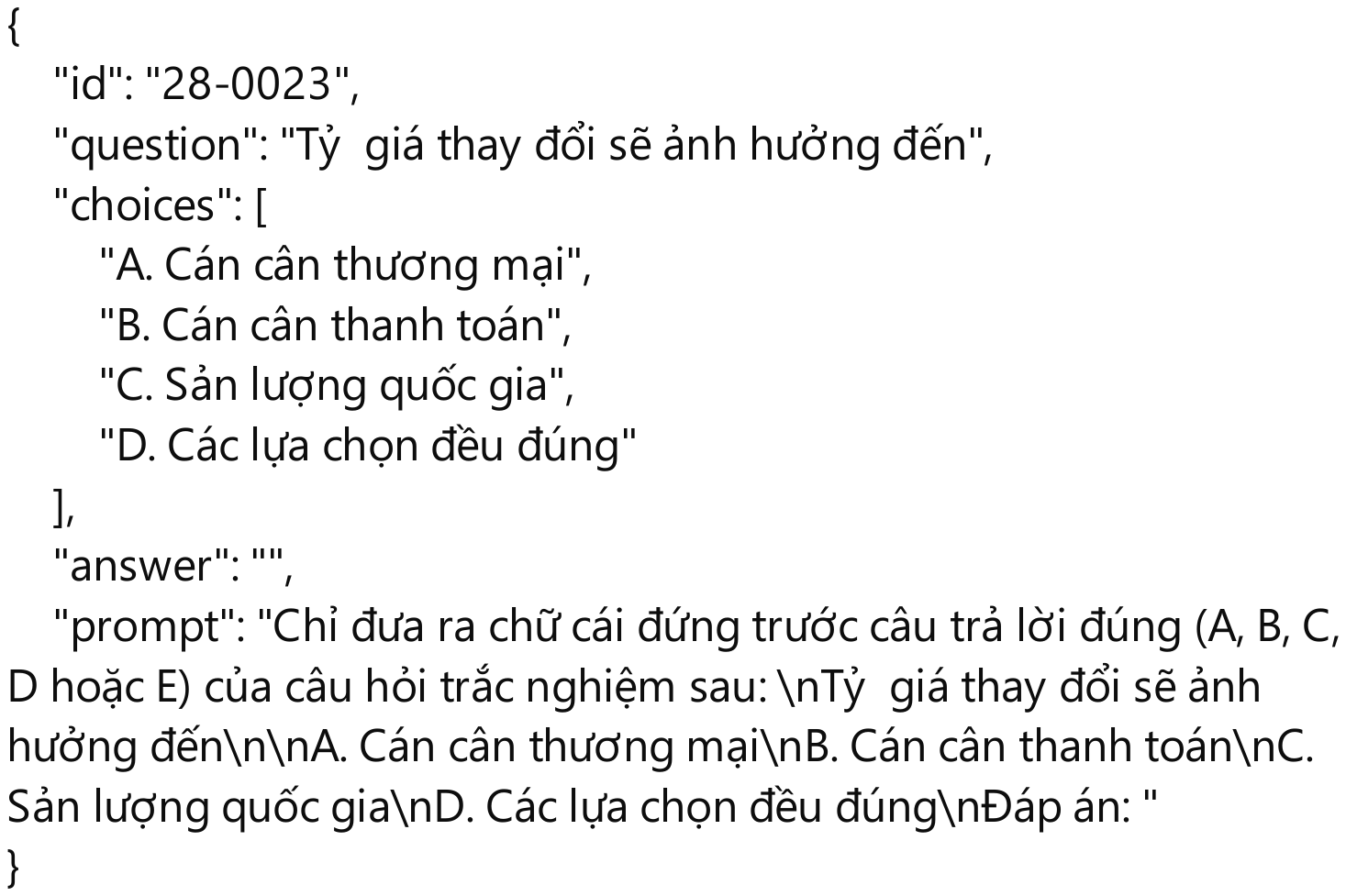}
    \caption{Example of VMLU} 
    \label{fig:yourlabel}
\end{figure}

\paragraph{Experimental Results}

\begin{table}[ht]
\centering
\begin{tabular}{l l l l l l l l l l}
\hline
\# & Model                        & Creator        & Access  & EvalDate & STEM   & SS & Hum & Others & Avg   \\
1  & GPT-4                        & OpenAI         & API     & 08/01/2024      & 63.84  & 71.78          & 66.14      & 60.37  & 65.53 \\
2  & gemini                       & Google         & API     & 30/01/2024      & 42.8   & 60.31          & 55.35      & 51.30  & 51.03 \\
3  & ChatGPT                      & OpenAI         & API     & 08/01/2024      & 43.24  & 51.67          & 46.96      & 46.32  & 46.33 \\
4  & ViGPT-1.6B-v1                & Vin BigData    & Private & 08/01/2024      & 35.06  & 48.72          & 47.20      & 42.54  & 42.34 \\
5  & gemma-7b-it                  & Google         & Weight  & 22/02/2024      & 39.95  & 44.93          & 43.39      & 40.11  & 41.9  \\
6  & Qwen-7B                      & Alibaba Cloud  & Weight  & 08/01/2024      & 30.64  & 35.07          & 34.15      & 32.68  & 32.81 \\
7  & \textbf{vi-mistral-x*}               & \textbf{James}          & \textbf{TBD}     & \textbf{15/03/2024}      & \textbf{24.88}  & \textbf{34.08}          & \textbf{35.11}      & \textbf{29.26}  & \textbf{30.32} \\
8  & gemma-2b-it                  & Google         & Weight  & 22/02/2024      & 24.39  & 29.59          & 31.01      & 26.81  & 27.72 \\
9  & sealion7b                    & AI Singapore   & Weight  & 08/01/2024      & 26.28  & 28.57          & 27.66      & 27.34  & 26.73 \\
10 & bloom-1b7                    & BigScience     & Weight  & 08/01/2024      & 25.13  & 25.09          & 26.34      & 25.19  & 25.51 \\
\hline
\end{tabular}
\caption{Comparision of Pretrained Models on VMLU}
\label{table:models_evaluation}
\end{table}
\textit{The model ``vi-mistral-x*'' is currently under development. The shown results were obtained by evaluating a checkpoint at epoch 0.08.}

The comparison is shown in Table 4, and the detailed evaluation of ``vi-mistral-x'' is presented in Table 5.

\begin{table}[ht]
\centering
\caption{Detailed Evaluation of VI-Mistral-X* on VMLU}
\label{your-label-here}
\begin{tabular}{l|r}
\hline
\textbf{Category\_Subcategory} & \textbf{Score} \\
\hline
total & 30.32 \\
stem\_applied\_informatics & 39.44 \\
stem\_computer\_architecture & 31.11 \\
stem\_computer\_network & 34.64 \\
stem\_discrete\_mathematics & 23.64 \\
stem\_electrical\_engineering & 22.73 \\
stem\_elementary\_mathematics & 19.44 \\
stem\_elementary\_science & 55.00 \\
stem\_high\_school\_biology & 15.00 \\
stem\_high\_school\_chemistry & 22.78 \\
stem\_high\_school\_mathematics & 16.22 \\
stem\_high\_school\_physics & 23.33 \\
stem\_introduction\_to\_chemistry & 14.53 \\
stem\_introduction\_to\_physics & 23.12 \\
stem\_introduction\_to\_programming & 29.05 \\
stem\_metrology\_engineer & 22.70 \\
stem\_middle\_school\_biology & 31.18 \\
stem\_middle\_school\_chemistry & 18.33 \\
stem\_middle\_school\_mathematics & 17.59 \\
stem\_middle\_school\_physics & 21.67 \\
stem\_operating\_system & 30.56 \\
stem\_statistics\_and\_probability & 10.34 \\
stem\_total & 24.88 \\
other\_clinical\_pharmacology & 26.11 \\
other\_driving\_license\_certificate & 45.61 \\
other\_environmental\_engineering & 11.70 \\
other\_internal\_basic\_medicine & 34.50 \\
other\_preschool\_pedagogy & 34.31 \\
other\_tax\_accountant & 20.69 \\
other\_tax\_civil\_servant & 41.52 \\
other\_total & 29.26 \\
other\_accountant & 21.43 \\
other\_civil\_servant & 27.49 \\
humanity\_economic\_law & 29.81 \\
humanity\_education\_law & 33.13 \\
humanity\_elementary\_history & 49.72 \\
humanity\_high\_school\_history & 31.11 \\
humanity\_high\_school\_literature & 25.56 \\
humanity\_history\_of\_world\_civilization & 41.11 \\
humanity\_idealogical\_and\_moral\_cultivation & 49.44 \\
humanity\_introduction\_to\_laws & 39.68 \\
humanity\_introduction\_to\_vietnam\_culture & 28.33 \\
humanity\_logic & 18.97 \\
humanity\_middle\_school\_history & 37.78 \\
humanity\_middle\_school\_literature & 37.36 \\
humanity\_revolutionary\_policy\_of\_the\_vietnamese\_commununist\_part & 36.67 \\
humanity\_vietnamese\_language\_and\_literature & 17.24 \\
humanity\_total & 35.11 \\
humanity\_administrative\_law & 37.78 \\
humanity\_business\_law & 39.11 \\
humanity\_civil\_law & 41.11 \\
humanity\_criminal\_law & 38.04 \\
social\_science\_middle\_school\_geography & 27.21 \\
social\_science\_principles\_of\_marxism\_and\_leninism & 36.67 \\
social\_science\_sociology & 39.89 \\
social\_science\_business\_administration & 20.69 \\
social\_science\_high\_school\_civil\_education & 43.89 \\
social\_science\_high\_school\_geography & 33.33 \\
social\_science\_ho\_chi\_minh\_ideology & 41.34 \\
social\_science\_macroeconomics & 21.67 \\
social\_science\_microeconomics & 23.89 \\
social\_science\_middle\_school\_civil\_education & 52.25 \\
social\_science\_total & 34.08 \\
\hline
\end{tabular}
\end{table}

\subsection{Finetuned Model}

The following session is being updated.

\bibliographystyle{unsrtnat}
\bibliography{references}  






\end{document}